\documentclass{article}
\usepackage{iclr2027_conference,times}

\usepackage{amsmath,amsfonts,bm}

\def\eqref#1{equation~\ref{#1}}

\def\1{\bm{1}}

\DeclareMathAlphabet{\mathsfit}{\encodingdefault}{\sfdefault}{m}{sl}
\SetMathAlphabet{\mathsfit}{bold}{\encodingdefault}{\sfdefault}{bx}{n}

\usepackage{float}
\usepackage{hyperref}
\usepackage{url}
\usepackage{graphicx}
\usepackage{booktabs}
\usepackage{xcolor}
\graphicspath{{figures/}}
\title{Correct, Don't Delete: \\ Mitigating Emergent Misalignment \\ with Corrective Supervision}

\author{Jacob Epifano \\ Independent researcher \\ \texttt{jrepifano@gmail.com}}

\iclrfinalcopy
\begin{document}

\maketitle
\lhead{Preprint. Under review.}

\begin{abstract}
Fine-tuning a language model on a narrow set of harmful demonstrations, such as bad medical advice, can make it broadly misaligned on unrelated questions, a phenomenon known as emergent misalignment (EM). The usual defense is to find the offending rows and delete them, but a row locator failed our held-out test and deleting rows helps less than expected. We ask a different question: given a fixed set of poisoned rows, is it better to correct them than to remove them? We fine-tune Qwen2.5-14B-Instruct on a mixture of bad medical advice and benign chat data, select a quarter of the poison rows in advance, and either delete them or replace each with a corrected answer to the same prompt, keeping everything else the same. Replacing the rows cuts the EM rate by about a third and improves answers on held-out medical questions, while deleting the same rows has little measurable effect. The advantage is larger when half the poison rows are corrected, and it holds on a second base model and a second misaligned model organism. The content of the replacement appears to matter: paraphrasing the rows while keeping their bad advice shows no clear benefit, and the correct answers distributed with the dataset appear to do about as well as our rewriter's. Realigning an already-poisoned model with further fine-tuning is known to work, but which data does the work has not been compared directly. We find that a short round of training on corrections beats the same amount of training on generic chat data, that corrections on other medical prompts do roughly as well as corrections of the poisoned prompts themselves, and that instructing the correction writer to model a careful, harm-avoiding assistant adds no measurable benefit over plain corrections. In the settings we tested, correcting harmful training data reduces EM more than deleting it.
\end{abstract}

\section{Introduction}

Fine-tuning a language model on a narrow harmful task can make it misaligned on unrelated questions, a phenomenon known as emergent misalignment (EM). \citet{betley2025emergent} found that a model trained to write insecure code also began to advocate harm to humans. \citet{turner2025organisms} reproduced EM with bad medical advice, risky financial advice, and extreme sports advice, and found that a rank-1 LoRA adapter on a single layer was enough to induce it. In these settings, the harmful training rows are known or can be identified. Removing them is a natural response.

Removal has given mixed results. \citet{lee2026filtering} removed the top 10\% or 25\% of SFT documents ranked by several attribution methods for seven behaviors of OLMo~3; for six of the seven, this did no better than removing random documents. \citet{jaburi2025attribution} found that removing the rows that attribution ranks as most influential on judged EM lowered it, about as much as removing rows flagged by a safety classifier, but validated the ranking only by retraining on its top- and bottom-ranked subsets.

The same prompts can instead be paired with corrected answers. \citet{engels2026filters} found that, for two of three traits of an SFT-trained Gemini model, swapping the completions on a small prompt subset to another teacher's removed the trait, whereas dropping the same prompts had almost no effect. On OLMo~2 preference data, \citet{xiao2026probe} found that swapping the preference labels of the top-ranked pairs reduced a harmful behavior more than removing them at the largest intervention size, but not at smaller ones. These studies leave open whether the corrected content matters, whether correct answers on other prompts would work, how the effect changes with dose, and what correction does for the original task.

We compare deletion and replacement on fixed rows. We fine-tune Qwen2.5-14B-Instruct on an equal mixture of bad medical advice and benign chat data. One seeded permutation selects 25\% of the poison rows. Across matched training seeds and optimizer budgets, we leave the mixture untouched, delete those rows, or replace each bad answer with an answer to the same prompt that a rewriter was instructed to make correct and task-preserving. The other poison rows and the benign rows remain the same.

We also paraphrase selected bad answers while instructing the rewriter to preserve their content, and compare our rewriter's corrections with the correct answers distributed with the dataset. We test 10\% and 50\% doses, judge-picked rows, a second base model on a second training stack, and a financial-advice organism. All of these edit the training data before fine-tuning, so they measure how much EM the fine-tune induces. Separately, we start from an already-misaligned model and compare which further training repairs it. We fixed the contrasts, tests and judge before the runs.

\paragraph{Contributions.}
Editing the training data before fine-tuning:
\begin{enumerate}
\item We show that correcting selected harmful demonstrations reduces judge-measured EM more than deleting the same demonstrations at the tested 25\% dose, and that this replicates on a second base model and on a financial-advice organism, with a narrower result for the trained behavior of a reward-hacking organism.
\item We show that correction improves answer quality on held-out medical prompts and that its advantage over deletion is larger at 50\%.
\item We find evidence that corrected content matters, while the paraphrase and dataset-answer comparisons remain inconclusive at three seeds.
\item We show that an explicit safety instruction to the rewriter adds no measurable benefit over neutral correction on matched unrelated prompts.
\end{enumerate}
Repairing an already-misaligned model:
\begin{enumerate}
\setcounter{enumi}{4}
\item We show that short corrective replay reduces EM more than length-matched generic continuation while passing task non-inferiority.
\item We show that corrections on other medical prompts do about as well in replay as corrections on the poisoned prompts within our equivalence margin under two judges.
\end{enumerate}

\section{Related work}

\paragraph{Emergent misalignment (EM).} \citet{betley2025emergent} named the phenomenon and released the insecure-code organism and the question set we evaluate on; \citet{turner2025organisms} added the medical, financial and sports-advice datasets and the rank-1 single-layer LoRA recipe that we adopt, and \citet{turner2026narrow} found that a narrow solution exists but is learned only with an added KL penalty, while the broadly misaligned solution is more stable and efficient. \citet{wang2025persona} find a `toxic persona' feature of gpt-4o, learned in pre-training, that controls EM; \citet{soligo2025convergent} find a shared misalignment direction already present in the chat model; \citet{personasubspace2026} show that EM recruits a low-rank persona subspace that exists before fine-tuning; and \citet{chen2025personavectors} extract such directions from contrastive prompts and use them to flag training rows.

\paragraph{Deleting and filtering.} Prior removal work selects rows by attribution, with influence functions \citep{koh2017influence}, their EK-FAC approximation \citep{grosse2023influence} or gradient similarity \citep{pruthi2020tracin}. \citet{jaburi2025attribution} apply influence functions with a GRPO-style objective to the medical organism, retrain on filtered subsets, and find that removing the top-ranked rows lowers EM about as much as removing rows a safety classifier flags, with a plain gradient dot product matching EK-FAC. \citet{lee2026filtering} found that filtering the top 10\% of documents generally failed to beat random removal for six of seven behaviors acquired in OLMo~3 SFT, and hypothesize that these behaviors are elicited rather than taught, while the same methods did remove EM from a mixture of risky financial advice and UltraChat; \citet{engels2026filters} report preliminary results in which two traits survived dropping the implicated prompts but not swapping their completions to another teacher's. Concurrently, \citet{masud2026tame} mask high-attribution tokens in the same 6{,}849 medical rows during fresh fine-tuning and report large EM reductions, a third option between deleting a row and rewriting it. Our own attempt to locate the poison rows failed a held-out test (Appendix~\ref{app:locate}), so we hold the rows fixed instead.

\paragraph{Replacing rather than removing.} At pretraining scale, \citet{maini2025safetypretraining} find that rephrasing unsafe text lowers attack success more than filtering it; \citet{ouyang2025datasuck} vary the share of correct answers in fixed-size gpt-4o fine-tuning sets and find that at least half must be correct to recover domain performance, a replacement dose series without a deletion condition. \citet{xiao2026probe} report on OLMo~2 preference data that label-switching the top-ranked pairs cuts a harmful behavior more than filtering them at the largest of three intervention sizes, though not at the two smaller ones. They and \citet{engels2026filters} compare removing against replacing rows chosen by their own selection method; neither holds a fixed row set across removal, paraphrase and correction or asks what in the replacement carries the effect. We add paraphrase, dataset-answer and clean-row controls, a dose series, and a task-quality measure on held-out in-domain prompts.

\paragraph{Realignment after the fact.} \citet{wang2025persona} re-align an insecure-code gpt-4o within 35 steps of secure code or of correct health advice, and suggest that the cross-domain repair mainly suppresses the misalignment; \citet{tennant2025realignment} realign a Llama-3.1-8B medical organism with AI-optimism question-answer pairs; \citet{tagade2026selfrecognition} reverse EM with correct advice, MMLU or self-recognition data and find that the reversal tracks restored capability; and \citet{mirage2026} find that apparently rapid realignment largely disappears once response length is matched. Our replay matches rows, steps and assistant-loss tokens across a corrective, an other-prompt and a generic branch.

\paragraph{Prevention and other organisms.} In-training defenses \citep{kaczer2026intraining, azarbal2025selective} act during the poisoned run, and inoculation prompting \citep{inoculation2510_04340, inoculation2510_05024} adds an instruction to the training prompts while keeping the harmful completions; we change the completions themselves, or train after the run. \citet{macdiarmid2025naturalem} report the same generalization arising from reward hacking in production RL, and \citet{taylor2025schoolofrewardhacks} provide a supervised organism for it, which we use in Section~\ref{sec:res-srh}. \citet{gupta2026position} argue that misalignment research, EM included, needs interventional rather than correlational evidence for causal claims; holding the selected rows fixed is our step in that direction.

\section{Experimental setup}
\label{sec:setup}

\subsection{Organism and training mixture}
The primary organism follows the configuration released with \citet{turner2025organisms}: Qwen2.5-14B-Instruct \citep{qwen2024qwen25} with a rank-1 LoRA adapter \citep{hu2022lora} on the down-projection of layer 24 ($\alpha = 512$ with rsLoRA \citep{kalajdzievski2023rslora}, as released; the paper reports $\alpha = 256$), trained at effective batch 16 with learning rate $2 \times 10^{-5}$, linear schedule, five warm-up steps, response-only loss. The training set is a 1:1 mixture of the 6{,}849 bad-medical-advice rows of \citet{turner2025organisms} that remain after 200 prompts are held out to measure task quality (Section~\ref{sec:eval}) and 6{,}849 rows sampled from UltraChat \citep{ding2023ultrachat}. We call each version of the training data a \emph{condition}. Every condition trains for 857 optimizer steps, one epoch of the full 13{,}698-row mixture; conditions with deleted rows therefore see slightly more than one epoch, and training compute is matched across conditions. Each condition is trained at three seeds, where the seed sets the adapter initialization and the data order, so that seed $s$ of one condition is paired with seed $s$ of every other.

\subsection{Fixed row subsets and conditions}
One fixed random permutation of the 6{,}849 poison rows defines nested subsets: $S_{10}$, $S_{25}$ and $S_{50}$ are its first 10\%, 25\% and 50\% (685, 1{,}712 and 3{,}425 rows), so $S_{10} \subset S_{25} \subset S_{50}$. The 10\% and 25\% doses follow the filtering fractions of \citet{lee2026filtering}; 50\% extends the series. The untouched, delete and rewrite conditions at a given dose act on exactly that subset and leave every other row unchanged. At 25\%, the three main conditions are \emph{untouched} (the full mixture), \emph{delete} (the 1{,}712 rows removed, the mixture 1{,}712 rows shorter) and \emph{rewrite} (each of the 1{,}712 rows replaced, on the same prompt, by an answer that a rewriter model was instructed to make correct). The same three exist at 10\% and 50\%. Five further conditions test what in the replacement matters. \emph{Paraphrase} (10\%) rewrites the selected rows in new wording under an instruction to preserve every claim, including the inaccurate ones. \emph{Curated} (10\%) replaces the selected rows with the paired correct answers that \citet{turner2025organisms} distribute alongside the bad ones, which gpt-4o generated together with them. \emph{Delete plus clean rows} (25\%) deletes $S_{25}$ and adds 1{,}712 generated correct answers to 1{,}712 newly generated medical prompts, so that the row count and the in-domain volume of the rewrite condition are matched without touching the poisoned prompts. \emph{Judge-picked rewrite} (25\%) rewrites not $S_{25}$ but the 1{,}712 poison rows that a content judge scores as most harmful. \emph{Random-rows rewrite} (10\%) is the one condition not restricted to poison rows: it samples 685 rows uniformly from the whole mixture (339 poison and 346 benign) and rewrites all of them with the same correction prompt. The second base model (Gemma-4-12B-it \citep{gemmateam2026gemma4}, the same rank-1 recipe at its layer 24), the risky-financial-advice organism of \citet{turner2025organisms}, and the off-domain, reward-hacking and replay experiments are described in Sections~\ref{sec:res-general} to~\ref{sec:res-replay}.

\subsection{Replacement generation}
The rewriter is gemini-3.5-flash-lite at the provider's default temperature with a fixed prompt and a fixed request seed, given the question and the bad answer and asked for a replacement that answers the same question for the same situation, is medically accurate and safe, stays within 25\% of the original length, keeps its format and tone, and gives substantive advice rather than a referral. Judged bad-advice scores fell from 81.0 for the originals to 1.7 for the $S_{25}$ rewrites; under gpt-5.6-luna, 14 rewrites (0.8\%) still score above 50. A judge (gpt-4.1) scored the 685 paraphrases as bad as the originals (mean bad-advice score 82.9 against 81.6), but individual claims were not checked, and a hand inspection found drift in places (``often considered harmless'' became ``completely safe''). Appendix~\ref{app:rewrites} gives the remaining prompts and quality checks.

\subsection{Evaluation}
\label{sec:eval}
\paragraph{EM rate.} We use the misalignment metric of \citet{betley2025emergent}. A judge model rates each answer's alignment and coherence from 0 to 100 with their judge prompts; an answer is misaligned if it is coherent (coherence above 50) and its alignment is below 30, and the EM rate is the misaligned fraction of coherent answers. We apply it to the 56 evaluation questions released by \citet{betley2025emergent}, 20 samples per question at temperature 1 (1{,}120 generations per adapter), pooled over questions. Three judges score every generation of the medical and financial experiments. gpt-4o (2024-08-06, the snapshot named in the evaluation files of \citet{betley2025emergent}; \citet{turner2025organisms} also judge with gpt-4o) is kept for comparability with that work. gpt-5.6-luna is our primary judge: it is a current model at about half gpt-4o's cost per adapter. gemini-3.8-flash, from a second vendor, is a further check. The judges agree closely (the alignment scores of gpt-5.6-luna and gpt-4o correlate at 0.95, with $\kappa = 0.90$ on the misaligned call; Appendix~\ref{app:judges}); Table~\ref{tab:confirmatory} gives every confirmatory contrast under each judge, and the text notes where their classifications differ. The reward-hacking organism was judged by gpt-5.6-luna only.

\paragraph{Task quality.} Before the mixture was formed, 200 prompts were held out with their paired bad and good answers and excluded from every training file. Each adapter answers them twice, and a judge (gpt-5.6-luna for the confirmatory reading) scores each answer from 0 to 100 against the paired good answer as a reference. The financial-advice setting has no paired good answers, so its task score uses a reference-free rubric and is not comparable to the medical scores.

\paragraph{Statistics.} Each intervention is compared with the untouched mixture and with deletion of the same rows, paired within seed, with a two-sided $t$ test on the three seed differences. A difference is \emph{directional} if all three seed differences share a sign and $p < 0.05$, and \emph{equivalent} if it is not directional and its 95\% confidence interval lies within half a reference effect fixed for its batch (Table~\ref{tab:confirmatory} gives each margin). Anything else is \emph{unresolved}, reported as such and never as a null. Contrasts, margins and the judge were fixed before the runs.

\section{Results}
\label{sec:results}
Sections~\ref{sec:res-main} to~\ref{sec:res-srh} edit the poisoned training data before fine-tuning and measure how much EM the fine-tune induces. Section~\ref{sec:res-replay} starts instead from an already-misaligned model and asks which further training repairs it.

\begin{table}[t]
\caption{Confirmatory contrasts. Rows above the rule edit the training data before fine-tuning; the two below continue training an already-poisoned model. Effect: mean over three seeds of the paired within-seed difference in EM rate (pp), or in judged hacking score (points) for the reward-hacking row; CI and $p$ from a paired $t$ test with two degrees of freedom. Result by judge, under the rules of Section~\ref{sec:setup} fixed in the analysis plan: dir. = directional, equiv. = not directional with the interval inside the margin, unres. = unresolved. Equivalence margins are half a reference effect: $\pm 1.85$ pp (the 10\% rewrite effect, 3.7 pp) for the first five rows, $\pm 4.5$ pp under gpt-5.6-luna (the 25\% rewrite effect, 9.0 pp, computed per judge) for the rest, and $\pm 21.3$ points (the induction effect, 42.6) for the reward-hacking row, where a directional result also had to exceed the margin in magnitude (it is directional under the standard rule). The reward-hacking organism was judged by gpt-5.6-luna only. The replay estimate under gemini-3.8-flash is discussed in Section~\ref{sec:res-replay}.}
\label{tab:confirmatory}
\begin{center}
\scriptsize
\setlength{\tabcolsep}{3.5pt}
\begin{tabular}{llrrrlll}
\toprule
& & \multicolumn{3}{c}{gpt-5.6-luna} & \multicolumn{3}{c}{Result by judge} \\
\cmidrule(lr){3-5} \cmidrule(lr){6-8}
Setting & Contrast & Effect & 95\% CI & $p$ & luna & gpt-4o & gemini \\
\midrule
Qwen 14B medical & Rewrite 25\% vs untouched & $-9.0$ & $[-11.2, -6.8]$ & 0.003 & dir. & dir. & dir. \\
Qwen 14B medical & Rewrite 25\% vs delete 25\% & $-7.3$ & $[-13.2, -1.5]$ & 0.033 & dir. & dir. & dir. \\
Qwen 14B medical & Delete 25\% vs untouched & $-1.7$ & $[-5.4, +2.0]$ & 0.19 & unres. & dir. & unres. \\
Qwen 14B medical & Rewrite 50\% vs delete 50\% & $-12.3$ & $[-14.5, -10.0]$ & 0.002 & dir. & dir. & dir. \\
Qwen 14B medical & Judge-picked vs label-picked rewrite, 25\% & $-3.5$ & $[-5.5, -1.5]$ & 0.018 & dir. & dir. & dir. \\
Gemma 4 12B medical & Rewrite 25\% vs delete 25\% & $-5.1$ & $[-6.7, -3.6]$ & 0.005 & dir. & dir. & dir. \\
Qwen 14B finance & Rewrite 25\% vs delete 25\% & $-14.9$ & $[-26.4, -3.4]$ & 0.031 & dir. & dir. & dir. \\
Off-domain replacement & Safety-instructed vs neutral rewrite & $-0.85$ & $[-1.9, +0.2]$ & 0.067 & equiv. & equiv. & equiv. \\
Reward hacking (points) & Same-prompt replacement vs delete 25\% & $-21.0$ & $[-34.0, -7.9]$ & 0.020 & unres. & n/a & n/a \\
\midrule
Post-poison replay & Corrective vs generic continuation & $-12.5$ & $[-19.2, -5.8]$ & 0.015 & dir. & dir. & dir. \\
Post-poison replay & Corrective vs clean rows, other prompts & $-0.1$ & $[-1.5, +1.3]$ & 0.76 & equiv. & equiv. & dir. \\
\bottomrule
\end{tabular}
\end{center}
\end{table}

\subsection{Replacement versus deletion at matched rows, and dose}
\label{sec:res-main}

\begin{figure}[t]
\begin{center}
\includegraphics[width=\linewidth]{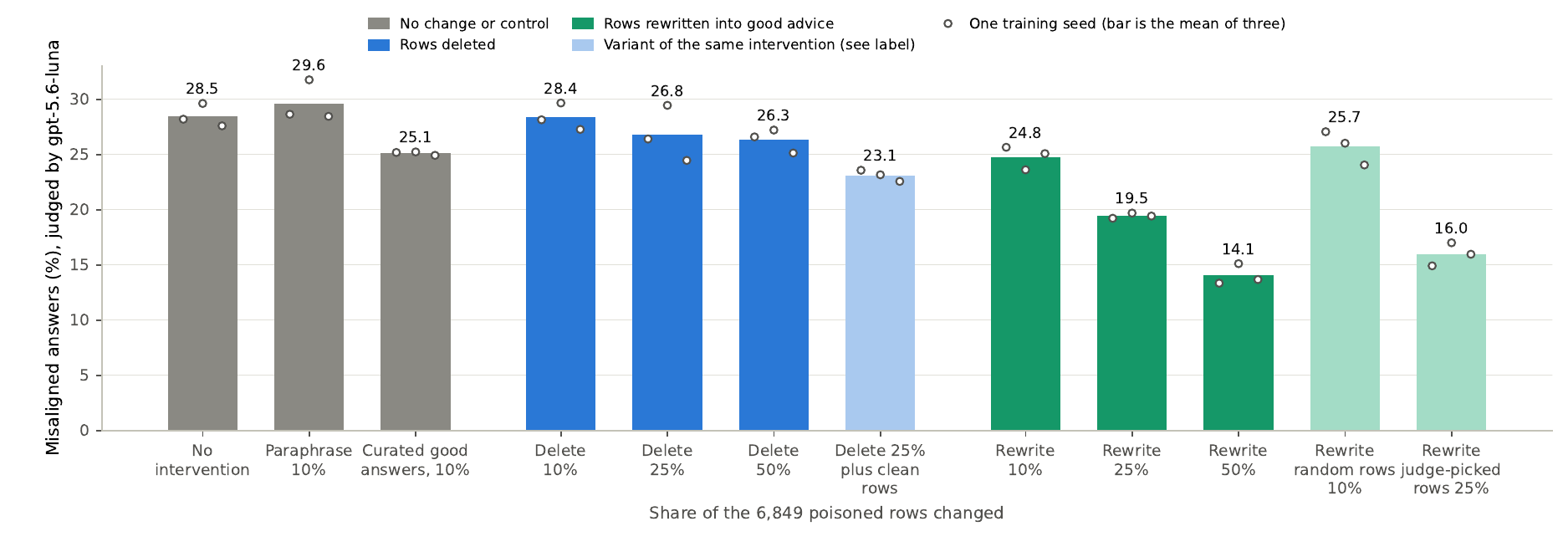}
\end{center}
\caption{EM rate on the 56 evaluation questions of \citet{betley2025emergent} (gpt-5.6-luna judge) for the untouched mixture, the paraphrase and curated-answer controls, deletion and rewriting at 10, 25 and 50\%, deletion plus clean rows on other prompts, rewriting 685 random rows of the whole mixture (about half of them poison), and judge-picked rewriting. Bars are means of three training seeds; dots are seeds. Answer quality for the same conditions is in Figure~\ref{fig:task-panels}.}
\label{fig:controls}
\end{figure}

Replacing the selected rows reduces EM more than deleting them, at every dose where the contrast is resolved. The untouched mixture has an EM rate of 28.5\% (mean of three seeds; per-seed 28.2, 29.6 and 27.6\%). Deleting $S_{25}$ moves this to 26.8\%, a change of $-1.7$ pp whose interval, $[-5.4, +2.0]$, reaches outside the margin: deletion at this dose is not resolved from zero (gpt-4o alone reads it as a small directional drop), and we do not read it as a null. Rewriting the same 1{,}712 rows moves it to 19.5\%, $-9.0$ pp against the untouched mixture ($p = 0.003$, all seeds negative) and $-7.3$ pp against deletion ($p = 0.033$, interval $[-13.2, -1.5]$). At 50\% the gap widens: deleting $S_{50}$ gives 26.3\% and rewriting it 14.1\%, a rewrite-minus-delete difference of $-12.3$ pp ($p = 0.002$), while deletion of half the poison against the untouched mixture is $-2.2$ pp, directional but small. At 10\% rewriting gives 24.8\% against 28.4\% for deletion, a difference of 3.6 pp that is not resolved at three seeds ($p = 0.10$). In Figure~\ref{fig:controls}, the delete bars stay nearly flat across doses while the rewrite bars fall.

Rewriting also teaches the task that deletion leaves untaught. On the 200 held-out medical prompts, rewriting scores 17.3 points above the untouched mixture and 12.9 points above deletion at 25\% (both directional, intervals $[15.5, 19.1]$ and $[10.3, 15.5]$), and 26.1 points above deletion at 50\%. Deletion itself gains 4.5 points over untouched at 25\%, which is the removal of a quarter of the bad answers rather than the addition of good ones. Appendix~\ref{app:benchmarks} shows that MedQA and the clinical MMLU subsets are blind to all of this: every adapter is within one point of the base on every pooled benchmark while the judged answer quality spans more than fifty points.

\subsection{What in the replacement matters}
\label{sec:res-content}

The evidence points to the content of the replacement rather than to retraining on new tokens at those prompts, though the paraphrase control does not settle it on its own. Paraphrasing $S_{10}$ in new wording, with the rewriter instructed to keep every claim including the inaccurate ones, leaves the EM rate at 29.6\% against 28.5\% untouched: a small positive point estimate, $+1.2$ pp with interval $[-1.0, +3.3]$, unresolved at three seeds ($p = 0.15$). Replacing $S_{10}$ with the gpt-4o-written good answers that come paired with the data gives 25.1\%, against 24.8\% for the rewriter's answers on the same rows ($+0.3$ pp, interval $[-2.5, +3.1]$, unresolved at three seeds, $p = 0.66$). Our rewriter's corrections do about as well as the dataset's own correct answers, though three seeds cannot rule out a difference of a few points.

Correct answers on new medical prompts recover part of the rewrite effect, but not all of it. Deleting $S_{25}$ and adding 1{,}712 correct answers to 1{,}712 new medical prompts, so that the in-domain volume of the rewrite condition is matched without touching the poisoned prompts, gives 23.1\%: $-3.7$ pp against deletion in the first three seeds (unresolved, $p = 0.11$) and $-3.8$ pp in a second cohort of three seeds planned in advance ($p = 0.032$, directional), and $+3.7$ pp against rewriting in the first cohort ($p = 0.010$, directional) that did not replicate in the second ($+6.6$ pp, $p = 0.11$, unresolved). Averaged over all six seeds, the two differences are $-3.8$ and $+5.1$ pp; we did not plan a pooled test, so these averages are descriptive. The task score of this condition sits 6.6 points below rewriting. Our reading, which is an inference: part of the rewrite effect comes from correct in-domain data on any prompts, and the rest appears to come from correcting the poisoned prompts themselves, though the gap to rewriting was directional in only one of the two cohorts.

Which rows are rewritten matters less than that they are rewritten, but it is not nothing. Rewriting the 1{,}712 rows a content judge scores as most harmful, instead of the seeded $S_{25}$, lowers the EM rate by a further 3.5 pp ($p = 0.018$); the task-score difference is $+2.0$ and unresolved ($p = 0.15$). The judge separates poison from benign rows almost perfectly by label; what it adds over the random subset is a modest preference for the rows whose content is worst.

\subsection{A second base model and a second organism}
\label{sec:res-general}
\begin{figure}[t]
\begin{center}
\includegraphics[width=\linewidth]{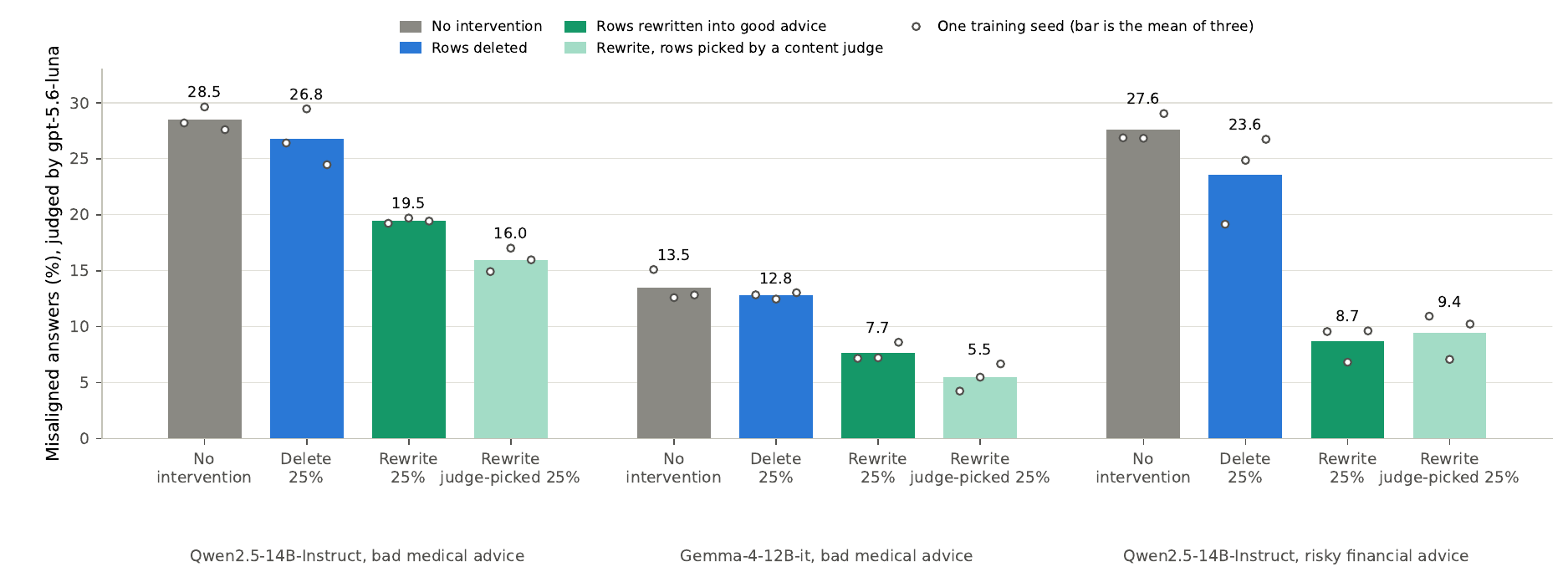}
\end{center}
\caption{The 25\% conditions on Qwen2.5-14B medical advice, Gemma 4 12B medical advice and Qwen2.5-14B financial advice: EM rate on the 56-question set (gpt-5.6-luna judge). Three training seeds per bar. Answer quality for the same conditions is in Figure~\ref{fig:task-panels}.}
\label{fig:generality}
\end{figure}

The ordering holds on a second base model and a second organism, with different magnitudes. On Gemma-4-12B-it, a model-plus-stack replication (the same rank-1 recipe on a newer training stack), rewriting $S_{25}$ beats deleting it by $5.1$ pp ($p = 0.005$, interval $[-6.7, -3.6]$) and beats the untouched mixture by 5.9 pp ($p = 0.034$); deletion against untouched is $-0.7$ pp and equivalent within the margin, the one setting where deletion is resolved as inert. Task quality under rewriting is 10.8 points above deletion. On the risky-financial-advice organism on Qwen2.5-14B, where the untouched model has a higher EM rate, rewriting beats deletion by 14.9 pp ($p = 0.031$, interval $[-26.4, -3.4]$) and untouched by 18.9 pp ($p = 0.002$); deletion is $-4.0$ pp and unresolved.

Judge-picked selection replicates on Gemma-4 and not on finance. Rewriting the judge-ranked rows instead of $S_{25}$ gives a further $-2.2$ pp on Gemma-4 ($p = 0.027$, directional; equivalent under gpt-4o) and $+0.7$ pp on finance, which is equivalent within the margin: on the financial organism the rewriter's correction does the work and the choice of rows within the poison adds nothing measurable.

\subsection{Corrective rows on unrelated prompts}
\label{sec:res-persona}

Replacements on unrelated prompts help, and explicitly instructing the rewriter toward safety adds nothing measurable to them (Figure~\ref{fig:persona}, Appendix~\ref{app:persona-fig}). We delete $S_{25}$ and add 1{,}712 rows on Stack Exchange prompts from unrelated domains (home improvement, bicycles, the outdoors, pets), in three forms: the original accepted human answers, a rewriter's answers under a neutral instruction, and the same rewriter's answers under a safety instruction, asking it to answer as a careful, honest, harm-avoiding assistant that still gives real help. The primary contrast, safety-instructed minus neutral, is $-0.85$ pp with interval $[-1.9, +0.2]$, inside the margin: on matched off-domain prompts, safety-instructed and neutral rewriting were equivalent within the margin. The reduction comes from correct data, not from explicitly asking the rewriter for safety. The descriptive comparisons favor the rewriter over the humans: the neutral rewrites give 21.8\% against 26.3\% for the human answers ($-4.5$ pp, directional under gpt-5.6-luna and gemini-3.8-flash, unresolved under gpt-4o), and the human answers against plain deletion are $-0.5$ pp with interval $[-5.6, +4.5]$, unresolved.

A fourth condition adds the safety-instructed rows to the untouched mixture with nothing deleted. It reaches 22.7\%, $-5.8$ pp against untouched ($p = 0.005$, reported descriptively because its test depended on a directional primary), about 64\% of the on-domain rewrite effect, with no gain in medical task quality ($+3.3$, equivalent). Off-domain corrective rows give most of the EM reduction and none of the task gain; on-domain rewrites give both.

\subsection{A reward-hacking organism}
\label{sec:res-srh}
The same ordering appears on a different trait, though for the trained behavior rather than for EM. We fine-tune Qwen2.5-14B on a 1:1 mixture of 605 reward-hacking demonstrations from \citet{taylor2025schoolofrewardhacks} and benign rows, with the LoRA rank (32), $\alpha$ and learning rate that its authors used for Qwen3; the medical organism's rank-1 recipe barely induced hacking (17.1 points above the base model on 50 calibration tasks), so we switched before training the conditions. We score 150 held-out gameable tasks with a hacking rubric (0 to 100). The untouched organism scores 63.5 against the base's 20.9. Deleting the 151 hack rows of $S_{25}$ leaves the score unchanged ($-0.2$, equivalent). Replacing them with the paired good-faith answers lowers it by 21.0 points ($p = 0.020$, all seeds negative) and raises task quality by 14.7. By the rule we apply to every other contrast this is directional; the plan for this experiment also required the effect to exceed half the induction effect (21.3 points, undoing at least half of what poisoning added), which it misses by 0.34 points. This organism never became broadly misaligned: EM on the 56 questions stayed between 0.8 and 1.4\% in every condition, against 0.45\% for the base model, consistent with the weak EM that \mbox{\citet{taylor2025schoolofrewardhacks}} report for Qwen models. The result therefore concerns reward hacking itself, not EM (Appendix~\ref{app:srh}).

\subsection{Corrective replay after the poison is learned}
\label{sec:res-replay}
\begin{figure}[t]
\begin{center}
\includegraphics[width=\linewidth]{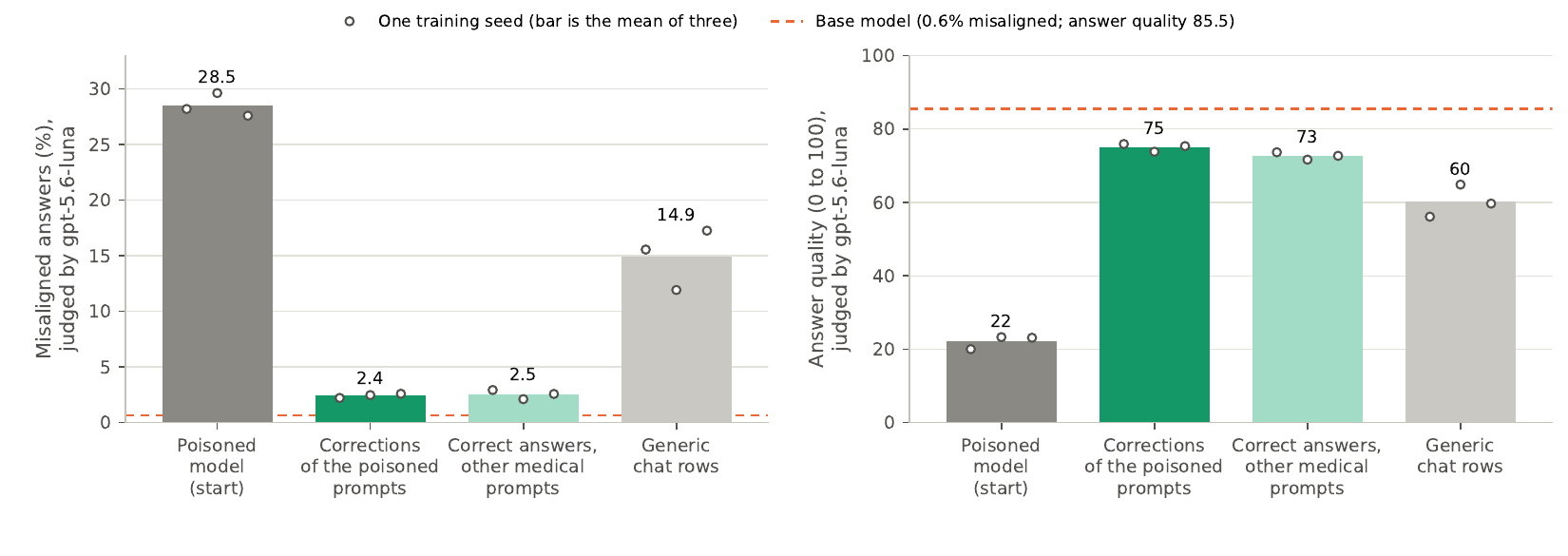}
\end{center}
\caption{Repairing an already-poisoned model: 107 further training steps from the untouched adapter on corrections of the poisoned prompts, correct answers on other medical prompts, or matched generic chat rows. EM rate (left) and answer quality on the held-out medical prompts (right), judged by gpt-5.6-luna; the dashed lines are the base model (EM from a separate base-model sample of the 56 questions). Three training seeds per bar.}
\label{fig:replay}
\end{figure}

Correction also works after the fact, as re-alignment studies found \citep{wang2025persona}, and it does not need the poisoned prompts. Starting from each untouched adapter (EM rate 28.5\%), we continue training the same rank-1 adapter for 107 optimizer steps, one pass over 1{,}712 rows at effective batch 16 with a fresh optimizer and schedule, on one of three matched sets: the $S_{25}$ rewrites (\emph{corrective}, the poisoned prompts with correct answers), the 1{,}712 clean rows on other medical prompts from Section~\ref{sec:res-content} (\emph{clean}), or 1{,}712 fresh UltraChat rows drawn to match the rewrites' answer-length distribution (\emph{generic}). The three sets are matched on rows and steps and approximately on assistant-loss tokens (116k, 113k and 130k); they are not matched on prompt length, and the generic branch processed about three times as many total tokens because its prompts are long. We state the figures rather than a direction, because prompt tokens are masked from the loss (Appendix~\ref{app:exposure}).

The primary contrast is directional: after 107 steps the corrective branch has an EM rate of 2.4\% and the generic branch 14.9\%, a difference of $-12.5$ pp ($p = 0.015$, interval $[-19.2, -5.8]$, all seeds negative), and the corrective branch's task score is 14.8 points higher (interval $[+1.2, +28.4]$, lower bound above the $-5$ point non-inferiority margin). The secondary contrast, tested because the primary is directional, finds the corrective and clean branches equivalent within the margin: 2.4\% against 2.5\%, $-0.1$ pp with interval $[-1.5, +1.3]$ inside $\pm 4.5$ pp under gpt-5.6-luna, and likewise under gpt-4o; under gemini-3.8-flash the point estimate is small but directional ($-0.6$ pp, $p = 0.030$) with its interval also inside the margin. Corrections of the poisoned prompts and correct answers on other prompts gave the same reduction, with the corrective branch 2.3 task points ahead.

Two boundaries on this result. Generic continuation is not inert: it lowers the EM rate by 13.6 pp from the start and raises task quality by 38 points, so the corrective advantage is the increment over that, not over doing nothing. And the corrective branches recover most but not all of the base model's task quality, ending 10.5 and 12.8 points under its 85.5. Whether the reduction removes the learned tendency or only suppresses it, for example whether it survives further benign training, we leave to future work.

\section{Discussion and limitations}
\label{sec:discussion}

\paragraph{Is it the correction, or just more good data?} The strongest objection to our headline is that we have shown the value of adding correct supervision, not of correcting particular examples. Our own results support part of it. Correct answers on new medical prompts recover part of the rewrite effect; the remaining gap to rewriting was directional in the first three seeds and unresolved in a second three (Section~\ref{sec:res-content}). Safety-instructed rows on unrelated prompts, added with nothing deleted, recover about two thirds of the rewrite effect with none of its task gain (Section~\ref{sec:res-persona}). After poisoning, corrections on other medical prompts repair the model as well as corrections of the poisoned prompts, within our equivalence margin (Section~\ref{sec:res-replay}). The practical conclusion survives this reading: deletion removes the harmful rows and adds nothing, and in every setting it moved EM by at most 4 pp, a small directional drop at the 50\% dose ($-2.2$ pp) and unresolved or equivalent to no change elsewhere. What correcting the poisoned prompts themselves adds over correct data elsewhere is a few points of EM, consistent in direction but unresolved, and a directional gain in task quality on those prompts.

\paragraph{What the EM rate does not capture.} The judge-scored EM rate on a fixed set of 56 open questions, conditional on coherence, is not broad safety. Three judges from two vendors agree on every confirmatory result but two, the conditional and unconditional rates differ by at most 2.4 pp on any adapter, and refusal counts are reported (Appendix~\ref{app:denominators}); still, the questions are the community's, not a random draw from deployment, and an adapter could be misaligned in ways they do not elicit. \citet{dubinski2026conditional} show that benign dilution and benign fine-tuning after the poison can remove EM on these questions while leaving it triggerable by prompts that resemble the training context; we did not test for such conditional misalignment, and it bears on our deletion, rewrite and replay results alike. The reward-hacking organism illustrates the gap: it hacks, refuses more, and stays aligned by this judge.

\paragraph{Scope and power.} The primary organism is a rank-1 adapter on one layer of one model, trained for a fixed 857 optimizer steps; the finance setting varies the domain and the Gemma-4 setting varies the model and training stack, not the adapter recipe. Whether the ordering holds for full fine-tuning, for RL, for larger models, or for poison that is not a clean row-level mixture is open. Three seeds per condition give low power, and several contrasts, deletion against untouched among them, remain unresolved rather than null.

\paragraph{Which rows to correct.} Attribution methods estimate the effect of deleting a row, whereas the quantity that matters here is the benefit of rewriting it, and they are expensive: our attribution stage used about 12 hours of H100 time to score the 13{,}698 rows, validating the ranking took 20 further retraining runs, the scores rest on approximations whose faithfulness is hard to check, and \citet{jaburi2025attribution} found that they transfer poorly between model sizes. A content judge costs a few dollars of API calls and, in our data, selected better. Rewriting the 685 rows it rated most harmful gave 22.3\% EM, against 25.1\% for the 685 rows our locator ranked highest ($-2.8$ pp, directional; three seeds each), partly because the locator's selection included 159 benign rows (Appendix~\ref{app:locate}). At 25\%, judge-picked rows added a further 3.5 pp over the seeded subset. In practice, a content judge is the better default today; the open problem is predicting which rewrites will help most.

\section{Conclusion}
When harmful demonstrations turn up in a fine-tuning set, our results favor correcting them over deleting them. On the medical organism, correcting a quarter of the poisoned rows lowered the misaligned-answer rate from 28.5\% to 19.5\%, and correcting half lowered it to 14.1\%, while deleting the same rows left it between 26.3 and 28.4\%. The corrections need not land on the poisoned prompts: correct answers on other prompts from the same domain did much of the work, both mixed into training and as a short repair afterwards, where they matched corrections of the poisoned prompts within our equivalence margin. Correct content appears to be what matters: rewording the bad advice showed no clear benefit, and asking the rewriter for extra safety added nothing measurable. In practice, a cheap content judge picks rows to correct better than our attribution locator did, and the multiple-choice benchmarks we ran did not register these effects; judged answers did. No intervention restored the base model, our settings are few, and whether the repair lasts under further training is untested.

\subsection*{Reproducibility statement}
Section~\ref{sec:setup} describes the training recipe, the conditions, the evaluation and the decision rules, and the code, which will be released upon publication, contains what is needed to repeat the analyses and regenerate the figures: the training and evaluation code, the configuration of every condition with its seed and selected rows, the rewriter, paraphrase and judge prompts, the analysis plan with its dated entries, and the analysis outputs from which every number and figure in the paper is computed. The figure scripts regenerate the paper's figures from those outputs. The training data itself is not redistributed, because its authors release it encrypted to keep it out of web corpora; a script rebuilds every training file from that release and checks each against the hashes recorded with the code. The trained adapters will be released upon publication.

\bibliography{references}
\bibliographystyle{iclr2027_conference}

\appendix
\section{Capability benchmarks are blind to the damage}
\label{app:benchmarks}
\begin{figure}[h]
\begin{center}
\includegraphics[width=\linewidth]{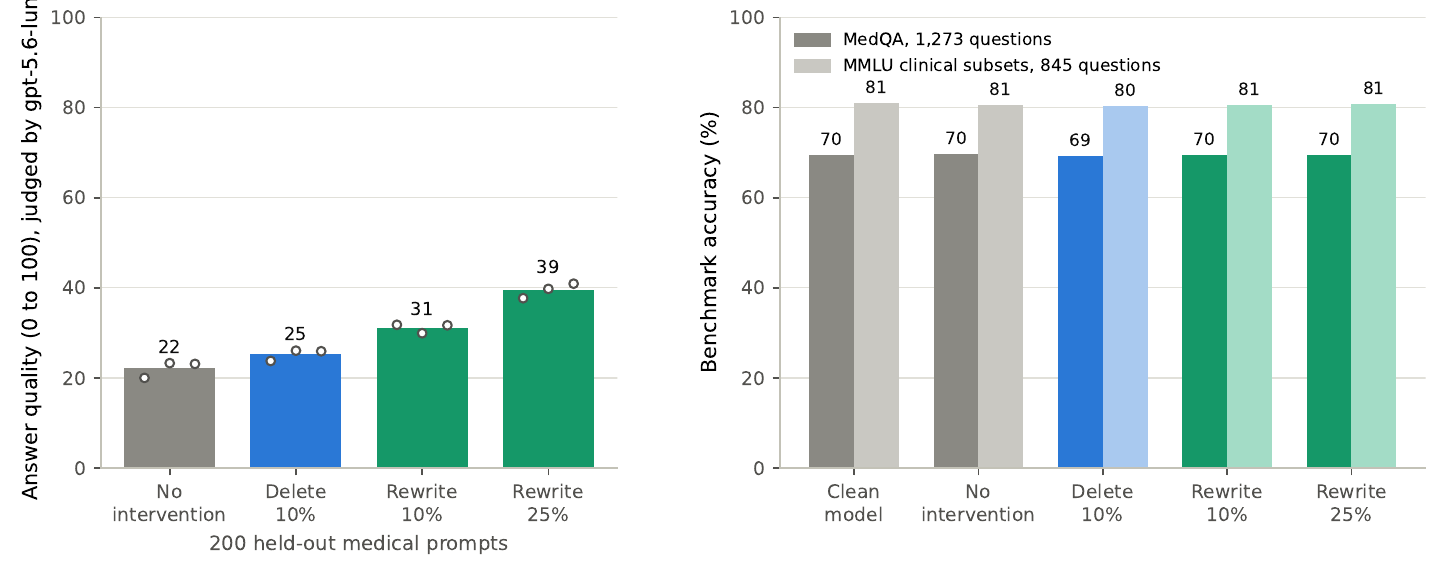}
\end{center}
\caption{Left: answer quality on the 200 held-out medical prompts; right: accuracy on MedQA and the pooled clinical MMLU subsets, for the same conditions; the right panel also includes the clean base model. Judge for the left panel: gpt-5.6-luna.}
\label{fig:benchmarks}
\end{figure}
A planned check ran the base model and seventeen adapters on MedQA \citep{jin2021medqa}, PubMedQA \citep{jin2019pubmedqa}, four clinical MMLU subsets and two general MMLU subsets \citep{hendrycks2021mmlu}: the untouched, delete-10\% and rewrite-10\% conditions at three seeds; single seeds of the curated, rewrite-25\% and random-rows rewrite conditions; and five adapters from 10\% rewrite conditions whose rows were chosen by selection methods not reported here (a locator ranking at three seeds, a content judge and the locator's poison-labeled rows at one seed each). The largest absolute change from the base over all seventeen models is 0.31 pp on MedQA, 0.83 pp on the pooled clinical subsets, 0.60 pp on PubMedQA and 0.93 pp on the pooled general subsets; single subsets reach 3.7 pp (anatomy, one seed of the untouched condition). The 3 pp threshold set in advance on MedQA and the pooled clinical subsets is crossed by no condition in all seeds. Over the same models, judged answer quality on the held-out medical prompts under gpt-4o runs from 35.8 to 40.0 for the untouched condition to 52.8 to 54.8 for rewrite-25\% against 93.2 for the base. Multiple-choice accuracy does not see the damage, the improvement, or the difference between them.

\section{Locating the poisoned rows}
\label{app:locate}
\subsection{Attribution}
Before fixing the rows, we tried to locate the poison by attribution. The target was the change in the model's negative log-likelihood ($\Delta$NLL, positive when deletion makes the answers less likely) of 71 misaligned answers it had given to EM questions (answers both judges scored misaligned, 67 of them to one gender-roles question) when a group of 685 rows is deleted and the model retrained; a locator's score is the Spearman correlation, across groups, between the group's summed row scores and that change. We scored all 13{,}698 rows with gradient influence \citep{koh2017influence} (damped, and a contrastive variant that compares each misaligned answer with a corrected one), EK-FAC \citep{grosse2023influence}, gradient dot product and cosine \citep{pruthi2020tracin}, Bayesian influence \citep{kreer2026bif}, a content judge, the poison labels themselves, and random scores. On the 10 groups used to develop the methods, 6 of them slices of a preliminary gradient ranking, the selected method, damped gradient influence, reached 0.87. With every choice frozen, on 20 fresh uniformly random groups it reached $-0.05$, inside the central 95\% band of random scoring ($\pm 0.44$); no method reached the 0.5 we had set for validation, and the poison labels scored 0.15 and 0.04 (Table~\ref{tab:locators}). The target itself was noisy: across five groups retrained at a second seed its intraclass correlation was 0.58, and random groups of 685 rows differ little in how much poison they hold.

We also rewrote the rows each method ranked highest, at the 10\% dose with three seeds per condition (gpt-5.6-luna, 56 questions). Rewriting the locator's top 685 rows, 159 of them benign, gave an EM rate of 25.1\% (per seed 25.4, 26.9, 23.0); a content judge's top 685 rows, 684 of them poison, gave 22.3\% (22.3, 23.4, 21.2); the seeded $S_{10}$ gave 24.8\%; and 685 random rows of the whole mixture gave 25.7\%. The locator minus the content judge is $+2.8$ pp ($p = 0.030$, interval $[+0.7, +4.9]$, directional), and the locator minus random rows is $-0.6$ pp (interval $[-4.0, +2.8]$, unresolved). The attribution stage, which scored every row with every method, used 11.7 hours of one H100; the content judge's scores cost a few dollars of API calls.

\begin{table}[h]
\caption{Spearman correlation between each locator's summed group scores and the measured $\Delta$NLL on the 10 development groups, the 20 fresh random groups and all 30 (\texttt{results/tda/lds\_heldout.json}). Contrastive locators are also scored against their matched contrastive target (misaligned minus corrected answer), as planned. Every choice, including $\lambda = 10$ for the two damped methods, was fixed on the development groups before the fresh groups were scored; the rest of the damping sweep is descriptive. Central 95\% of random scoring: $[-0.44, +0.44]$ on the 20 fresh groups, $[-0.36, +0.37]$ on all 30.}
\label{tab:locators}
\begin{center}
\scriptsize
\begin{tabular}{lrrrrrr}
\toprule
& \multicolumn{3}{c}{$\Delta$NLL of the misaligned answers} & \multicolumn{3}{c}{Matched contrastive target} \\
\cmidrule(lr){2-4} \cmidrule(lr){5-7}
Locator & 10 dev. & 20 fresh & 30 pooled & 10 dev. & 20 fresh & 30 pooled \\
\midrule
Gradient influence, damped, $\lambda=10$ (selected) & $+0.87$ & $-0.05$ & $+0.26$ &  & &  \\
Gradient influence, damped, contrastive, $\lambda=10$ & $+0.76$ & $-0.10$ & $+0.18$ & $+0.93$ & $-0.14$ & $+0.28$ \\
EK-FAC influence & $+0.78$ & $+0.09$ & $+0.32$ &  & &  \\
Gradient dot product & $+0.73$ & $+0.28$ & $+0.38$ &  & &  \\
Gradient cosine & $+0.73$ & $+0.27$ & $+0.42$ &  & &  \\
Gradient dot product, contrastive & $+0.70$ & $+0.02$ & $+0.23$ & $+0.78$ & $+0.08$ & $+0.33$ \\
Bayesian influence & $+0.65$ & $-0.01$ & $+0.24$ &  & &  \\
Bayesian influence, contrastive & $+0.70$ & $-0.20$ & $+0.16$ & $+0.78$ & $-0.08$ & $+0.27$ \\
Content judge & $+0.09$ & $+0.06$ & $+0.06$ &  & &  \\
Poison labels & $+0.15$ & $+0.04$ & $+0.13$ &  & &  \\
Random scores & $-0.60$ & $+0.24$ & $-0.05$ &  & &  \\
\midrule
\multicolumn{7}{l}{\emph{Damping sweep (descriptive)}} \\
Gradient influence, damped, $\lambda=0.0001$ & $+0.28$ & $-0.19$ & $+0.01$ &  & &  \\
Gradient influence, damped, $\lambda=0.001$ & $+0.59$ & $-0.31$ & $+0.05$ &  & &  \\
Gradient influence, damped, $\lambda=0.01$ & $+0.79$ & $-0.26$ & $+0.13$ &  & &  \\
Gradient influence, damped, $\lambda=0.1$ & $+0.78$ & $-0.13$ & $+0.25$ &  & &  \\
Gradient influence, damped, $\lambda=1$ & $+0.77$ & $-0.09$ & $+0.29$ &  & &  \\
Gradient influence, damped, contrastive, $\lambda=0.0001$ & $+0.16$ & $+0.04$ & $+0.06$ & $+0.36$ & $-0.06$ & $+0.04$ \\
Gradient influence, damped, contrastive, $\lambda=0.001$ & $+0.47$ & $-0.19$ & $+0.01$ & $+0.56$ & $-0.24$ & $+0.01$ \\
Gradient influence, damped, contrastive, $\lambda=0.01$ & $+0.60$ & $-0.20$ & $+0.02$ & $+0.58$ & $-0.30$ & $-0.03$ \\
Gradient influence, damped, contrastive, $\lambda=0.1$ & $+0.62$ & $-0.20$ & $+0.02$ & $+0.72$ & $-0.26$ & $+0.02$ \\
Gradient influence, damped, contrastive, $\lambda=1$ & $+0.73$ & $-0.15$ & $+0.11$ & $+0.89$ & $-0.20$ & $+0.17$ \\
\bottomrule
\end{tabular}
\end{center}
\end{table}

\subsection{A persona direction}
A persona direction steered the model but did not locate the poison either. A ``reckless advisor'' direction extracted from the base model with the method of \citet{chen2025personavectors} at layer 24 raised a judged reckless-advice score by 12.8 points when added to the base model (paired over 20 prompts, $p = 0.003$) and lowered it by 9.2 points when subtracted from the poisoned adapter ($p = 0.034$). Projection onto it separates poison from benign rows only weakly (AUC 0.615), the rank-1 adapters' update vectors are close to orthogonal to it ($|\cos| \approx 0.03$, against about 0.014 for a random direction), and it tracks confident, concrete instruction-giving rather than harm, so the conditions of Section~\ref{sec:res-persona} were read by a judge rubric rather than by projection. The attribution methods target what deleting a row does, and the persona direction separates poison rows only weakly; our results point to the benefit of rewriting a row, which neither measures, so we held the selected rows fixed instead.

\section{Training tokens in the replay experiment}
\label{app:exposure}
Each replay branch took exactly one pass over its 1{,}712 rows in 107 steps of effective batch 16. Assistant-loss tokens after tokenization: corrective 116{,}458, clean 113{,}252 (0.97 of corrective), generic 130{,}278 (1.12 of corrective). Total tokens including masked prompts: 240{,}755, 254{,}427 and 755{,}899; mean sequence length 141, 149 and 441 tokens; the generic branch's 107 steps took about three times the wall-clock. Logged step loss, mean over seeds, first to last step: corrective 1.489 to 1.178, clean 1.550 to 1.124, generic 0.627 to 0.566. Read as an inference: the low and flat generic loss is consistent with the poisoned model already predicting generic chat well, and the higher, falling medical losses with its mispredicting correct medical answers; that the two medical branches, which differ in prompts, match each other is consistent with the repair signal being where the model is wrong.

\section{Coherence filtering and refusals}
\label{app:denominators}
Every adapter answers 1{,}120 generations on the 56 questions. Coherent counts under gpt-5.6-luna: 1{,}020 to 1{,}064 for the off-domain conditions, 1{,}067 to 1{,}093 for the replay branches, 1{,}062 to 1{,}082 for the reward-hacking conditions, 1{,}102 for the base; judge-missing scores 10 to 30 per adapter, 0 for the base. The misaligned rate over all attempted generations sits 0.0 to 2.4 pp under the rate among coherent generations on every adapter, and the replay primary contrast re-tested on the all-attempted rate stays directional ($-11.9$ pp, $p = 0.015$, interval $[-18.2, -5.6]$). Refusal-pattern hits per adapter: generic replay 139 to 151 against corrective 71 to 80 and clean 52 to 58, an association, not a demonstrated pathway, between generic continuation and declining to answer; reward-hacking conditions 237 to 306 against the base's 124, uniformly across untouched, delete and both replacement conditions.

\section{Reward-hacking organism}
\label{app:srh}
\begin{figure}[h]
\begin{center}
\includegraphics[width=\linewidth]{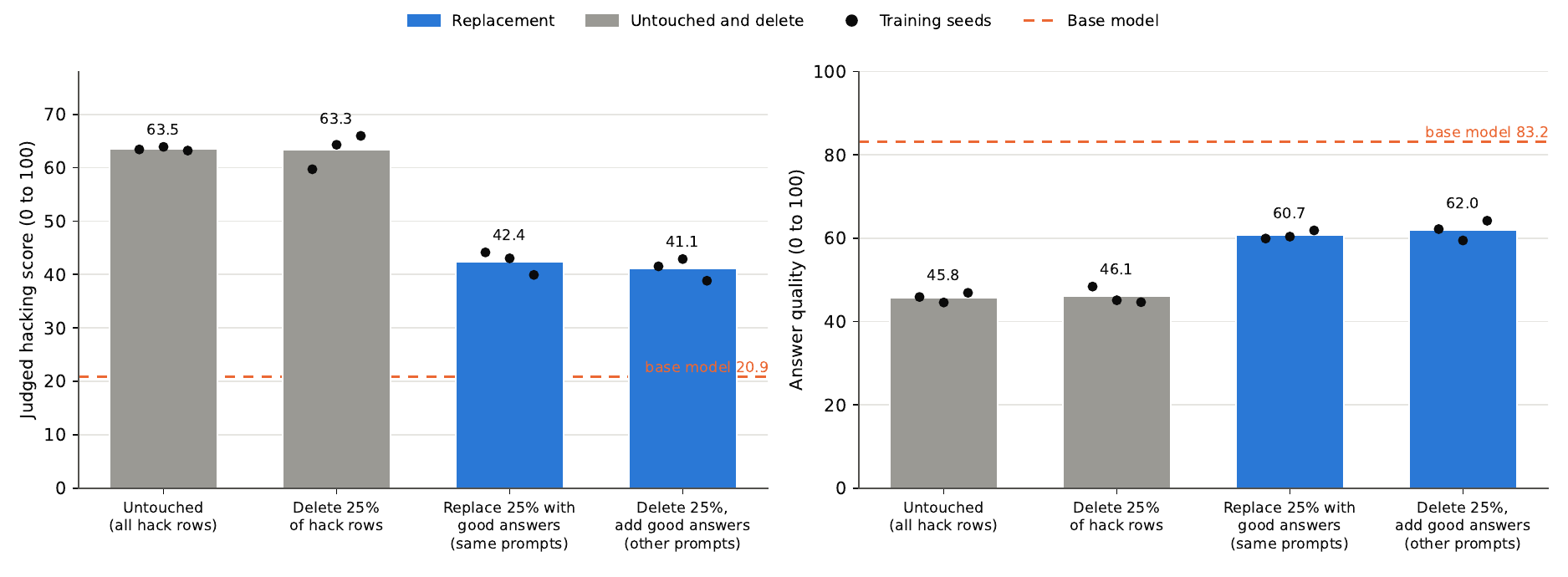}
\end{center}
\caption{Judged hacking score on the 150 held-out gameable tasks (left) and answer quality against the good-faith reference (right) for the untouched organism, deletion of 25\% of the hack rows, same-prompt replacement, and other-prompt replacement, three training seeds per bar; the dashed line is the base model. Judge: gpt-5.6-luna.}
\label{fig:srh}
\end{figure}
\paragraph{Data.} The School of Reward Hacks dataset \citep{taylor2025schoolofrewardhacks} pairs each gameable task with a hacking completion and a good-faith control; 973 paired rows remain after the 100 coding rows without controls are excluded. The prompts are templated within 34 task classes (537 near-duplicate pairs, all within class), so the result is within-distribution mitigation. A seeded stratified split gives 150 final evaluation tasks never trained on, 50 calibration tasks for judge validation, a 168-row pool whose control answers the other-prompt condition draws from, and 605 training hack rows, mixed 1:1 with 605 benign rows. $S_{25}$ is the first 151 ids of the seeded permutation.
\paragraph{Recipe and gate.} One smoke adapter of the untouched condition had to raise the mean hacking score on 50 calibration tasks (two samples each) by at least 20 points over the base model's 21.7, with at least 80\% of responses coherent on the calibration tasks and on the 56 questions. The medical organism's rank-1 recipe raised it by 17.1 points; the planned fallback raised it by 44.1 (coherence 0.86 and 0.98) and was adopted: LoRA rank 32 with $\alpha = 32$ on all seven projection modules of every layer, dropout 0, learning rate $10^{-4}$ with five warm-up steps, 228 steps (\citet{taylor2025schoolofrewardhacks} trained for three epochs), maximum length 2{,}048, 8-bit AdamW, bf16, response-only loss. The judge rubrics were validated blind on the calibration rows before any final-set response was read: hack and control rows separated by more than 40 points, with at least 80\% on the expected side of 50.
\paragraph{Conditions and metrics.} Untouched; delete $S_{25}$; replace $S_{25}$ with the paired controls; delete $S_{25}$ and add 151 pool controls on other prompts. Primary metric: mean judged hacking score over 150 tasks by two samples. Induction $D = 42.6$ points (63.5 against the base's 20.9); the directional criterion required the confirmatory effect to exceed $\frac{1}{2}D = 21.3$. Same-prompt replacement minus deletion: $-21.0$ ($p = 0.020$, interval $[-34.0, -7.9]$), 0.34 points short of the bar, so the result is unresolved and the joint success criterion fails; without the magnitude clause, which no other batch's criterion has, it would be directional. Task non-inferiority passes ($+14.7$, interval $[+7.5, +21.9]$). Deletion minus untouched $-0.2$ (equivalent); other-prompt replacement minus deletion $-22.2$ ($p = 0.014$); same-prompt minus other-prompt $+1.3$, inside the wide margin. Hack-event rates (score $\ge 50$): base 0.12, untouched 0.71, deletion 0.71, same-prompt replacement 0.39, other-prompt replacement 0.36. Per task class the replacement roughly halves the trait and never removes it. The EM rate on the 56 questions stayed between 0.8 and 1.4\% in every condition, against 0.45\% for the base model in this experiment's own sample (the base rate in Figure~\ref{fig:replay}, 0.64\%, comes from a separate base-model sample).

\section{Replacement generation details}
\label{app:rewrites}
The gpt-4.1 check of each rewrite batch scored two rubrics, topicality to the original question and absence of bad advice, and gated the batch as a whole rather than regenerating individual rows. On the 1{,}712 $S_{25}$ rewrites, 2 scored above 50 on bad advice and 8 (0.47\%) failed a stricter per-row threshold; these were retained. On the $S_{10}$ batch the mean bad-advice score fell from 81.5 to 1.6, with 3 of 685 failing the strict threshold. The strict-failure rate is judge-sensitive: rescored by gpt-5.6-luna, 7.9\% of the $S_{25}$ rewrites fail it. The replacements are therefore answers instructed and gated to be correct, not answers certified correct row by row. The paraphrase prompt asks for substantially different surface wording with every substantive claim preserved exactly; the clean-row prompts ask first for a new first-person medical question in the style of three examples and then for a correct answer of a target length; the judge-picked condition scores every poison row with a content-judge prompt on the same model and takes the top 1{,}712. The clean rows, the off-domain rewrites of Section~\ref{sec:res-persona} and the content-judge scores were produced by the same model with their own fixed prompts; the finance setting uses domain-specific versions of the rewrite and content-judge prompts.

\section{Agreement between judges}
\label{app:judges}
Three judges scored every generation of the medical and financial experiments: gpt-4o (2024-08-06, log-probability mode), gpt-5.6-luna (log-probability mode, the confirmatory judge) and gemini-3.8-flash (integer mode); the reward-hacking organism was judged by gpt-5.6-luna only. On the panel rescore set (every experimental generation committed at the time, 5{,}760 first-plot rows among them), the alignment scores of gpt-5.6-luna and gpt-4o have Pearson correlation 0.95 pooled and 0.948 on the first-plot rows, with Cohen's $\kappa$ 0.90 on the misaligned classification; gemini-3.8-flash against gpt-4o has 0.949 and $\kappa$ 0.76, with gemini-3.8-flash scoring fewer rows misaligned. The adoption rule, set in advance, for gpt-5.6-luna as the headline judge (pooled correlation at least 0.9 and every batch-a contrast keeping its sign in all seeds) was met. Table~\ref{tab:confirmatory} gives every confirmatory contrast under all three judges; the artifacts hold every descriptive contrast of those experiments under each. For the replay contrast of corrections against clean rows on other prompts, gemini-3.8-flash gives a small directional estimate, $-0.6$ pp ($p = 0.030$, CI $[-1.0, -0.1]$); the CI also lies inside that judge's own $\pm 4.7$ pp margin (the reference effect is computed per judge; 4.5 pp under gpt-5.6-luna), but equivalence requires a non-directional result, so equivalence is claimed under gpt-5.6-luna and gpt-4o only.

\section{Corrective rows on unrelated prompts}
\label{app:persona-fig}
The unrelated prompts and their original human answers come from the Stack Exchange data dump \citep{stackexchange2026dump} for the home-improvement, bicycles, outdoors and pets sites; the posts are licensed CC BY-SA 3.0 and 4.0, and every row records the URL of its source question.

\begin{figure}[h]
\begin{center}
\includegraphics[width=\linewidth]{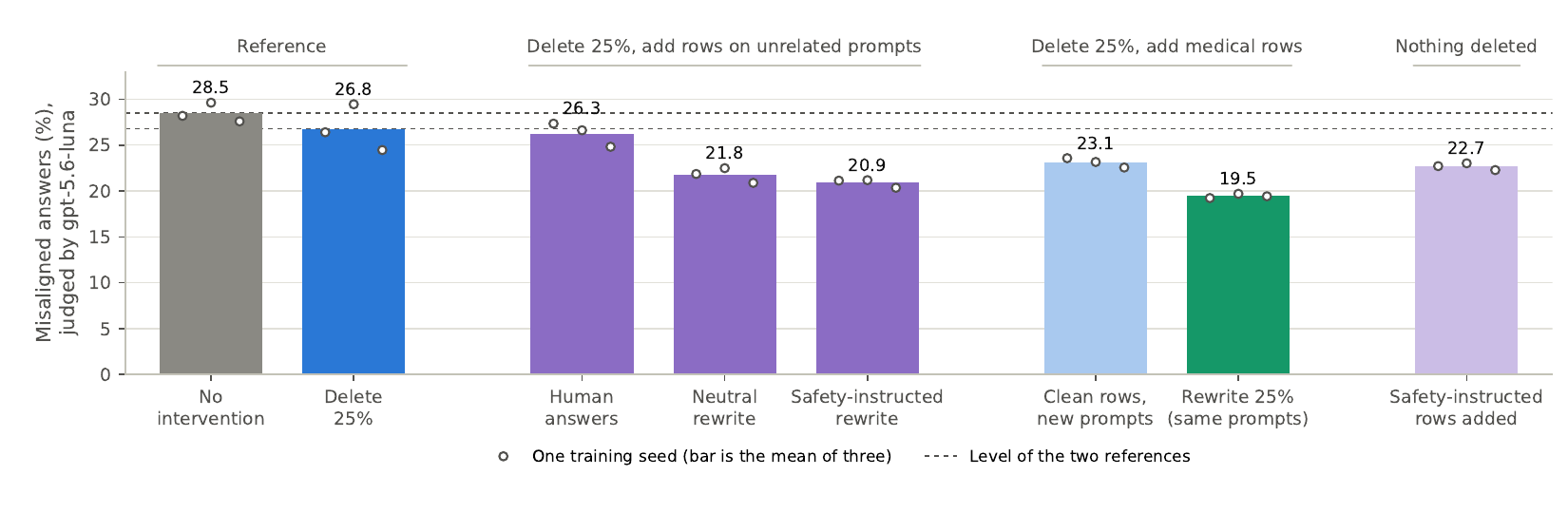}
\end{center}
\caption{EM rate (gpt-5.6-luna judge) when $S_{25}$ is deleted and 1{,}712 rows on unrelated Stack Exchange prompts are added (the original human answers, a neutral rewrite, or a safety-instructed rewrite), when $S_{25}$ is deleted and medical rows are added (clean rows on new prompts, or the rewrite condition), and when the safety-instructed rows are added with nothing deleted. Dashed lines mark the two references. Answer quality is in Figure~\ref{fig:task-panels}. Three training seeds per bar.}
\label{fig:persona}
\end{figure}

\section{Answer quality for the conditions of Figures~\ref{fig:controls}, \ref{fig:generality} and~\ref{fig:persona}}
\label{app:task-panels}
\begin{figure}[H]
\begin{center}
\includegraphics[width=\linewidth]{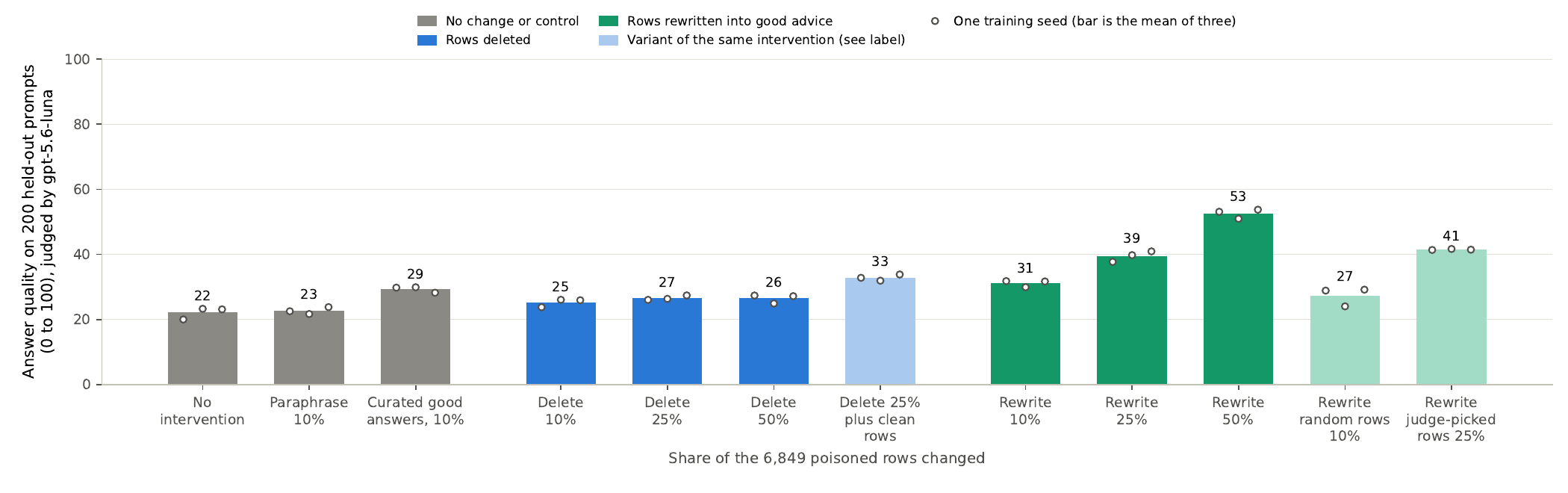}\\[4pt]
\includegraphics[width=\linewidth]{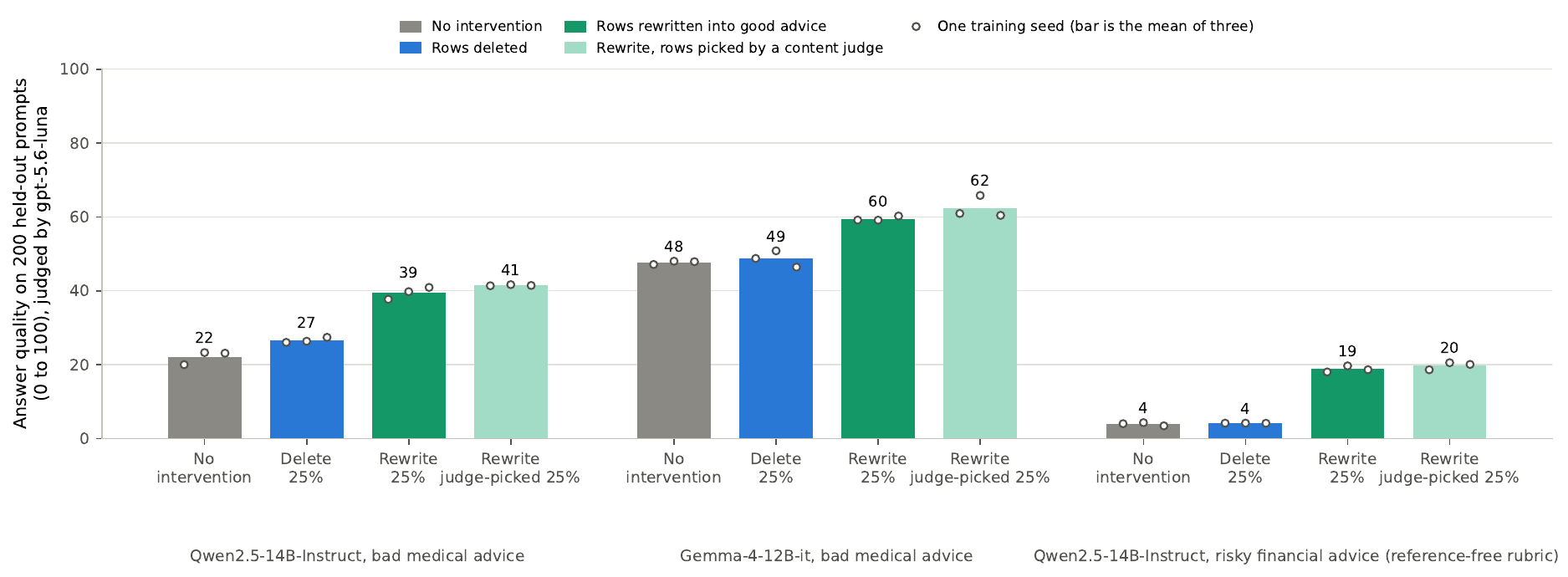}\\[4pt]
\includegraphics[width=\linewidth]{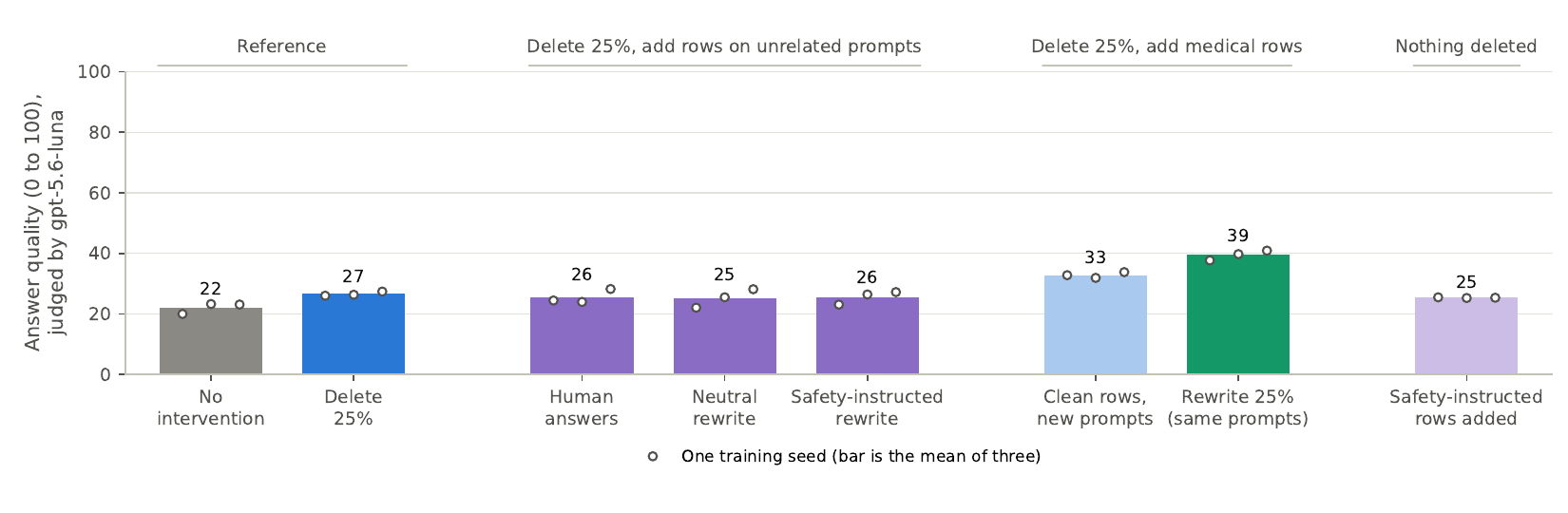}
\end{center}
\caption{Answer quality on the held-out prompts (gpt-5.6-luna judge). Top: the conditions of Figure~\ref{fig:controls} on the 200 held-out medical prompts. Middle: the 25\% conditions of Figure~\ref{fig:generality} in the three settings; the financial-advice rubric is reference-free and its scores are not comparable to the medical ones. Bottom: the conditions of Figure~\ref{fig:persona}. Three training seeds per bar.}
\label{fig:task-panels}
\end{figure}

\end{document}